\documentclass[english]{article}
\usepackage{spconf, amsmath, graphicx}
\usepackage{amssymb}
\usepackage[unicode=true,pdfusetitle,bookmarks=true,bookmarksnumbered=false,bookmarksopen=false,breaklinks=false,pdfborder={0 0 1},backref=false,colorlinks=false]{hyperref}
\usepackage{subfig}
\usepackage[backend=bibtex, firstinits=false, style=ieee]{biblatex}
\title{When does Bone Suppression and Lung Field Segmentation \\ Improve Chest X-Ray Disease Classification?}
\name{Ivo M. Baltruschat$^{1, 2, 3, 4}$
	\qquad Leonhard Steinmeister$^{1, 3}$
	\qquad Harald Ittrich$^{1}$
	\qquad Gerhard Adam$^{1}$
}
\secondlinename{Hannes Nickisch$^{4}$
	\qquad Axel Saalbach$^{4}$
	\qquad Jens von Berg$^{4}$
	\qquad Michael Grass$^{4}$
	\qquad Tobias Knopp$^{1, 2}$
}
\address{
	$^{1}$ Department for Diagnostic and Interventional Radiology and Nuclear Medicine, \\University Medical Center Hamburg-Eppendorf, Hamburg, Germany \\
	$^{2}$ Institute for Biomedical Imaging, Hamburg University of Technology, Hamburg, Germany \\
	$^{3}$ DAISYLabs, Forschungszentrum Medizintechnik Hamburg, Hamburg, Germany \\
	$^{4}$ Philips Research, Hamburg, Germany
}
\begin{document}

	\maketitle
	
	\begin{abstract}
		Chest radiography is the most common clinical examination type. To improve the quality of patient care and to reduce workload, methods for automatic pathology classification have been developed. In this contribution we investigate the usefulness of two advanced image pre-processing techniques, initially developed for image reading by radiologists, for the performance of Deep Learning methods. First, we use bone suppression, an algorithm to artificially remove the rib cage. Secondly, we employ an automatic lung field detection to crop the image to the lung area. Furthermore, we consider the combination of both in the context of an ensemble approach.
		
		In a five-times re-sampling scheme, we use Receiver Operating Characteristic (ROC) statistics to evaluate the effect of the pre-processing approaches. Using a Convolutional Neural Network (CNN), optimized for X-ray analysis, we achieve a good performance with respect to all pathologies on average. Superior results are obtained for selected pathologies when using pre-processing, i.e. for \textit{mass} the area under the ROC curve increased by $9.95\%$. The ensemble with pre-processed trained models yields the best overall results. 
		
	\end{abstract}
	
	\begin{keywords}
		neural network, chest X-ray, bone suppression, lung field detection, turnaround time
	\end{keywords}
	
	\section{Introduction}
	\label{sec:intro}
	With the already high and most likely increasing demand, chest radiography is today the most common examination type in radiology departments \cite{CQC2018}. As reported by \cite{beardmore2016radiography}, the average report turnaround time for plain X-ray is about 34 hours while $74\%$ have a turnaround time less than $24$ hours. In case of critical findings such as \textit{pneumothorax} or \textit{pleural effusion} the integration of automated detection systems in the clinical work-flow could have a substantial impact on the quality of care.

	Recent developments in pathology classification focused mainly on specific aspects of Deep Learning (e.g. in terms of novel network architectures). Early on, Shin \textit{et al.} \cite{Shin2016} demonstrated that a Convolutional Neural Network (CNN) combined with a recurrent part can be applied for image captioning in chest X-rays. The increased availability of annotated chest X-ray datasets like ChestX-ray14 \cite{Wang2017} helped to accelerate the progress in the field of pathology classification, detection and localization. 
	
	In this rapidly evolving field, Li \textit{et al.} \cite{Li2018} presented a unified network architecture for pathology classification and localization, while only limited annotation is needed for the localization.
	Cai \textit{et al.} \cite{Cai2018IterativeAM} proposed an attention mining method for disease localization which works without localization annotation. 
	In the work of Wang \textit{et al.} \cite{wang2018tienet}, a classification and reporting method -- leveraging the radiologist report in addition to the image -- was presented.

	In this context, only very simple pre-processing steps have been employed. Motivated by prior work in the computer vision domain this includes predominantly intensity normalization as well as a re-scaling of the image to the model size. Contrary, over the last years, several methods have been developed for supporting radiologists in the diagnostic process. Two well known techniques are bone suppression and lung field detection \cite{von2016novel, von2016decomposing}; the former artificially removes the rib cage facilitating the detection of small appearing pathologies and the later standardizes viewing appearance. In multiple studies, the usefulness of such image processing methods for different diseases was shown \cite{Li2012}. An obvious question arises: do bone suppression and lung field detection have the same beneficial effect on disease classification with CNNs? 
	
	Toward this end, we investigate how bone suppression and lung field detection can be exploited as a pre-processing step for a CNN. 
	
	In a methodology comparable way to \cite{Gordienko2018}, we apply pre-processing in three different scenarios. First, processing each image with bone suppression. Secondly, cropping the images to detected lung fields and finally, combining both processing steps. However, different to \cite{Gordienko2018}, we use lung field detection to crop the images to the important area, whereas Gordienko \textit{et al.} kept the image size equal and just set regions (not belonging to the lung fields) to zero. We believe cropping can increase the CNN performance as it increases the effective spatial resolution for the CNN. Furthermore, we propose a novel ensemble architecture to leverage the complimentary information in the different images, similar to a radiologists work-flow. Furthermore, in order to allow for a detailed assessment of the impact for specific pathologies, two expert radiologists, annotated, the public Indiana dataset (Open-I) with respect to eight findings.

	\begin{figure}[!ht]
		
		\centering
		\subfloat["Normal"]{\includegraphics[width=0.24\textwidth]{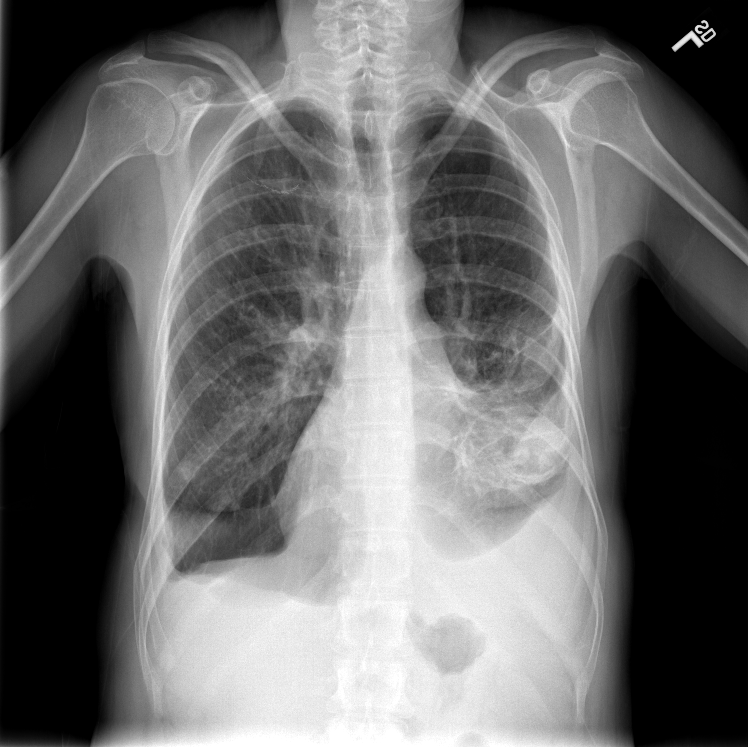}}\hfill
		\subfloat["Bone suppressed" \label{fig:img-BS}]{\includegraphics[width=0.24\textwidth]{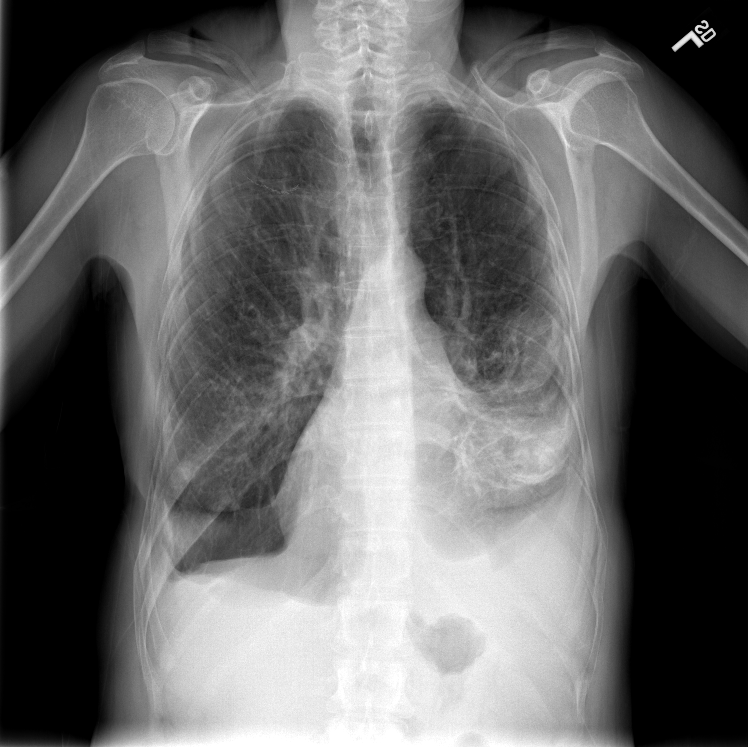}}\hfill
		\subfloat["Lung field cropped" \label{fig:img-Crop}]{\includegraphics[width=0.24\textwidth]{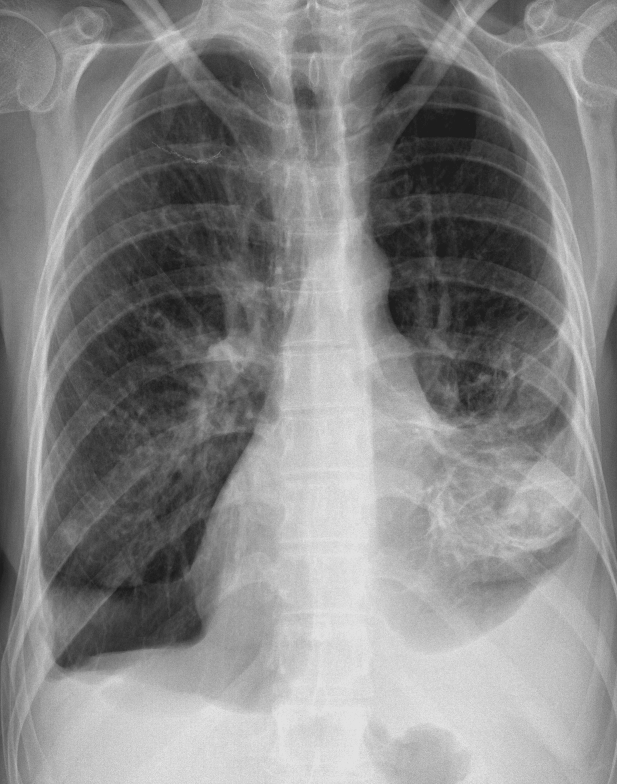}}\hfill
		\subfloat["Combination" \label{fig:img-combinied}]{\includegraphics[width=0.24\textwidth]{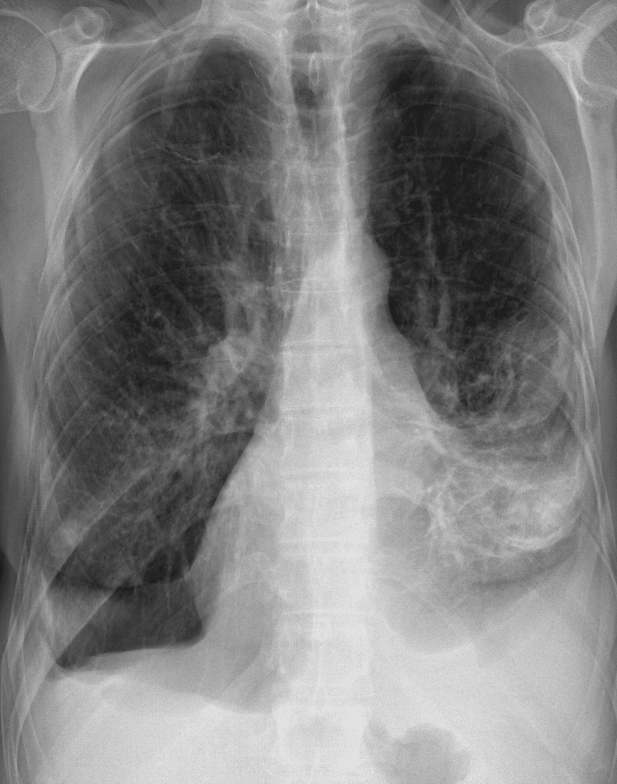}}
		
		\caption[Titel des Bildes]{One example image (a) of the Indiana chest X-ray dataset from Open-I. The dataset consists of 3125 frontal and lateral images from 3125 patients. We annotated all images with up to eight findings. The pre-processed images are show in (b)-(d). (b) is the bone suppressed image, (c) is cropped to the lung field, and (d) illustrates the combination out of (b) and (c).}
		\label{fig:indiana}
		
	\end{figure}
	
	\section{Methods}
	\label{sec:methods}
	Following the method and training setup in \cite{baltruschat2018comparison}, we pre-trained a ResNet-50 architecture with a larger input size of $448\times448$ on ChestX-ray14. Compared to different network architectures and training strategies, the obtained model achieved the highest average AUC value in our previous experiments. Due to the focus on eight specific pathologies, we replaced the last dense layer of the converged model with a new dense layer having eight outputs and a sigmoid activation function. Furthermore, we applied a fine-tuning step in order to adapt the model to the new image domain.
	
	\subsection{Bone Suppression}
	\label{sec:boneSup}
	In the original Indiana images we suppress the bones (ribs and clavicles) using a method from \cite{von2016novel, von2016decomposing}. The method preserves the remaining details originally overlaid with the bones (see Fig. \ref{fig:img-BS}). In the reported reader study, the AUC for the detection and localization of lung nodules increased for experienced human readers when using bone suppression images. Machine learning may potentially also benefit from suppressing some normal anatomy, which is to be tested here.
	
	\subsection{Lung Fields}
	\label{sec:lungField}
	Lung fields are segmented using a foveal CNN as described in \cite{brosch2018foveal}. It is trained by semi-automatically annotated lung fields and applied to the Indiana images. After the initial lung field detection, we apply post-processing steps to determine the final crop area. First, we identified all connected regions and computed a bounding box around the two largest region. Thereafter, we added a small border of $100$ pixel to the top/bottom and to the left/right. Each image is cropped to its individual bounding box as pre-processing step (see Fig. \ref{fig:img-Crop}). Lung field cropping has two beneficial aspects. First, it reduces the amount of information loss due to down scaling and secondly, it is a geometric image normalization. We also consider a combination of both -- bone suppression and lung field cropping (Fig. \ref{fig:img-combinied}).

	\subsection{Ensemble}
	\label{sec:ensamble}
	In many applications combining different predictors can lead to improved classification results, which is known as \textit{ensemble} forming \cite{Hansen1990, krogh1995neural}. Ensembling can be done in several ways and with any number of predictors. To determine whether the combination of several predictors could improve results, the Pearson correlation coefficient can be used. Ensembling predictors with high correlation coefficient will likely not improve results a lot compared to predictors with lower correlation. 
	
	Methods for ensemble generation include averaging and majority voting as well as machine learning algorithm like Support Vector Machines (SVMs). Since an ensemble approach will typically outperform an individual model, we compare not only individual models (trained with a specific pre-processing) to an ensemble trained on different images. Instead, we compare also an ensemble with models trained on images without pre-processing to a ensemble with pre-processed trained models. In order to limit the complexity of the experimental setup, we focus on averaging approach.
	
	\section{Indiana Dataset}
	\label{sec:data}
	The Indiana dataset from Open-I contains 3996 studies with DICOM images \cite{Demner-Fushman2015}. In a first step, we created a revised dataset, by removing studies with no associated images or labels (i.e. the reference annotation). Next, studies that lacked either frontal or lateral acquisition were removed. The final dataset consists of 3125 studies. Two expert radiologists from our department reviewed all cases and diagnosed, which findings are present using the frontal as well as the lateral acquisition. As shown in Table \ref{tab:dataset}, we have selected eight different findings for annotation: pleural effusion, infiltrate, congestion, atelectasis, pneumothorax, cardiomegaly, mass, and foreign object.
	
	Intra-observer variability is common in chest X-rays. Thus, after an individual assessment of the images, all disagreements were discussed and a final consensus annotation was found. Table \ref{tab:dataset} shows the distribution of each finding. All classes except pneumothorax have more than 100 positive cases, whereas the class pneumothorax only has eleven positive cases. In our final evaluation, we do report results but will not discuss them for pneumothorax because of the low number of positive cases.
	
	We re-sampled 5 times from the entire Indiana dataset for an assessment of the generalization performance\cite{Molinaro2005}. Each time, we split the data into 70\% training and 30\% testing. We calculated the average loss over all re-samples to estimate the best point for generalization. Finally, our results are calculated for each split on the test set and averaged afterwards. 
	
	\begin{table}
		
		\centering
		\caption{Indiana dataset statistic overview for all eight findings.}
		\label{tab:dataset}
		\begin{tabular}{l c c c}
			\noalign{\smallskip}
			Finding & True & False & Prevalence [\%] \\
			\noalign{\smallskip}
			\hline
			\noalign{\smallskip}
			pleural effusion & 147 & 2978 & 4.7\\
			infiltrate & 152 & 2973 & 4.9\\
			congestion & 170 & 2955 & 5.4\\
			atelectasis & 212 & 2913 & 6.8\\
			pneumothorax & 11 & 3114 & 0.4\\
			cardiomegaly & 529 & 2596 & 16.9\\
			mass & 447 & 2678 & 14.3\\
			foreign object & 1121 & 2004 & 35.9\\
		\end{tabular}
	\end{table} 
	
	\section{Experiments and Results}
	\label{sec:empirical}

	\noindent \textbf{Implementation:} Following the experimental setup in \cite{baltruschat2018comparison}, we employed an adapted ResNet-50, which is tailored to the X-ray domain. After replacing the dense layer, the model was fine-tuned using the Indiana dataset. For training, we sample various sized patches of the image with sizes between $80\%$ and $100\%$ of the image area. The patch aspect ratio is distributed evenly between $3:4$ and $4:3$. In addition, each image is randomly horizontal flipped and randomly rotated between $\pm7^{\circ}$. At testing, we resize images to $480 \times 480$ and use an averaged five crop (i.e. center and all four corners) evaluation. In all experiments, we use ADAM \cite{Kingma2015} as optimizer with default parameters for $\beta_1 = 0.9$ and $\beta_2 = 0.999$. The learning rate $lr$ is set to $lr = 0.005$. While training, we reduce the learning rate by a factor of 2 when the validation loss does not improve. We use a batch size of 15 and binary cross entropy as a loss function. The models are implemented in CNTK and trained on GTX 1080Ti GPUs.
	
	We perform six different experiments based on our proposed image pre-processing (Section \ref{sec:lungField} and \ref{sec:boneSup}). First, we train on normal images (i.e. no pre-processing is employed), bone suppressed images, lung cropped images, and on images combining both pre-processing steps. Secondly, we build an ensemble upon four normal trained models "EN-normal" as a baseline ensemble. Finally, we us our pre-processed trained models to build an ensemble "EN-pre-processed". 
	
	\begin{figure}
		
		\centering
		\subfloat["Normal trained"]{\includegraphics[width=0.33\textwidth]{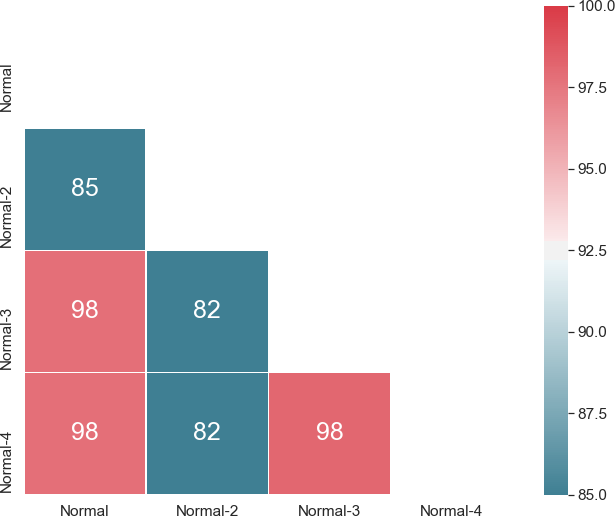}}\hfill
		\subfloat["Pre-processed trained"]{\includegraphics[width=0.33\textwidth]{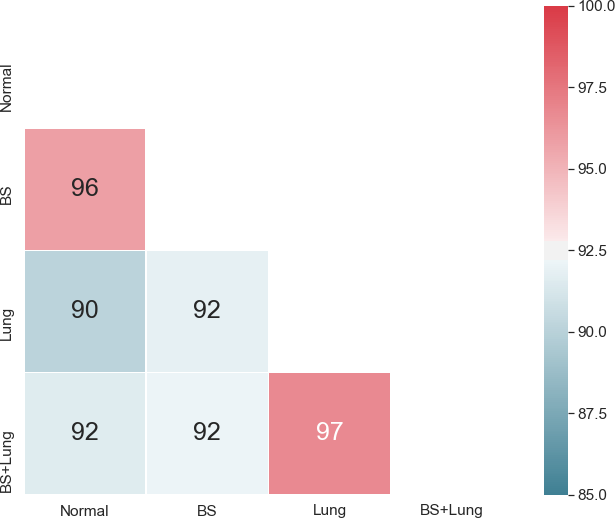}}\hfill

		\caption[Titel des Bildes]{Pearson correlation coefficient results for normal trained models (a) and models trained with pre-processed images (b). The correlation between normal models is higher than the models trained with pre-processed images. This is an indication, that an ensemble for the models in (b), result in an better result improvement.}
		\label{fig:rankCorr}
		\vspace{-0.5cm}
	\end{figure}
	\noindent \textbf{Results:} 
	\begin{table*}[t]
		
		\centering
		\caption{AUC result overview for all our experiments. In this table, we present averaged results over all 5 splits and the calculated standard deviation (SD) for each finding. Furthermore, the average (AVG) AUC over all findings is shown. We trained our model with four different input images. First, normal images. Secondly, "BS" means with bone suppressed images. Thirdly, "Lung" means with images cropped to lung fields. Fourthly, "BS+Lung" means with bone suppressed and cropped to lung fields. In addition, we formed an ensemble with models trained on normal images "EN-normal" and an ensemble with the models trained on pre-processed images "EN-pre-processed". Bold text emphasizes the overall highest AUC value. $^\bigstar$We excluded pneumothorax because of the low positive count.}
		\label{tab:results}
		\begin{tabular}{l c c c c c c}
			\noalign{\smallskip}
			Finding & Normal & BS & Lung & BS+Lung & EN-normal & EN-pre-processed  \\
			\noalign{\smallskip}
			\hline
			\noalign{\smallskip}
			pleural effusion& $.951 \pm .008$& $.948 \pm .009$& $.955 \pm .007$& $.955 \pm .009$& $\mathbf{.960} \pm .004$& $.957 \pm .007$\\
			infiltrate & $.936 \pm .012$& $.938 \pm .012$& $.939 \pm .007$& $.936 \pm .014$& $\mathbf{.944} \pm .010$& $.943 \pm .011$\\
			congestion & $.937 \pm .013$& $.932 \pm .015$& $.941 \pm .014$& $.938 \pm .014$& $.941 \pm .012$& $\mathbf{.946} \pm .013$\\
			atelectasis & $.905 \pm .020$& $.907 \pm .016$& $.917 \pm .017$& $.913 \pm .020$& $.905 \pm .020$& $\mathbf{.923} \pm .016$\\
			cardiomegaly & $.952 \pm .006$& $.950 \pm .006$& $.953 \pm .005$& $.952 \pm .003$& $.955 \pm .004$& $\mathbf{.959} \pm .003$\\
			mass & $.764 \pm .016$& $.766 \pm .016$& $.821 \pm .020$& $.840 \pm .011$& $.769 \pm .014$& $\mathbf{.837} \pm .014$\\
			foreign object & $.795 \pm .015$& $.815 \pm .013$& $.808 \pm .013$& $.805 \pm .015$& $.811 \pm .018$& $\mathbf{.821} \pm .015$\\
			pneumothorax & $.731 \pm .134$& $.789 \pm .104$& $.813 \pm .132$& $.794 \pm .128$& $.736 \pm .163$& $.792 \pm .142$\\
			\noalign{\smallskip}
			\hline
			\noalign{\smallskip}
			AVG $^\bigstar$ & $.891 \pm .013$ & $.894 \pm .012$ & $.905 \pm .012$ & $.906 \pm .012$ & $.898 \pm .012$ & $\mathbf{.912} \pm .011$ 
		\end{tabular}
	\end{table*}
	To compare our experiments to each other, we calculate the area under the ROC curve (AUC). The shown AUC results are averaged over all re-sampling and presented with standard deviation (SD). For our ensemble experiment, we calculate the Pearson correlation coefficient between each normal trained model and our pre-processed trained models. 
	
	First, we look at our experiments with the different pre-processed images and the performance based on AUC. In all experiments, we note that five out of seven relevant classes have a high AUC of above $0.9$. Two of those five \textit{pleural effusion} and \textit{cardiomegaly} have even an AUC of above $0.95$. Only the class \textit{mass} and \textit{foreign object} have an AUC below $0.9$. Comparing the results of a model using bone suppression to the normal trained model, the AUC for \textit{foreign object} increased significant from $.795 \pm .015$ to $.815 \pm .013$ with respect to the reported standard deviation (SD). The model trained with lung cropping has in all classes a higher AUC and often a reduced SD compared to the baseline. But only for the class \textit{mass}, the AUC increased significantly from $.766 \pm .016$ to $.821 \pm .020$. We argue that the increased spatial resolution for lung cropped images helps the model to better detect small masses. This is in line with the observation of our radiologists, which reported an increased number of small masses. Combining both pre-processing steps results in the highest AUC for \textit{mass} and increases the AUC by $9.95\%$. We observe no significant changes for the other classes.
	
	Secondly, we build two ensembles: \textit{EN-normal} and \textit{EN-pre-processed}. EN-Normal refers to our ensemble of four models trained using images without pre-processing. Whereas EN-pre-processed is an ensemble with one normal, one BS, one lung cropping, and one combined model. In figure \ref{fig:rankCorr}, we report the Pearson correlation coefficient for the normal and pre-processed ensemble. As expected, the four normal models are already highly correlated (i.e. values around 96) except for one model which seems to converged to a different optimum. Comparing the Pearson correlation coefficients of the pre-processed trained models with the normal trained models, the coefficient are lower and only around 85. This indicates that a pre-processed ensemble can have a high impact on our results. We verify our hypothesis with the AUC results in Table \ref{tab:results}. The pre-processed ensemble increases the AUC in \textit{mass}, \textit{foreign object}, and \textit{atelectasis} significantly with respect to the reported SD, whereas the normal ensemble does not. Overall, the pre-processed ensemble yields in five of seven classes the best AUC results.
	
	\section{Conclusion}
	\label{sec:discuss}
	In this contribution we investigated the effect of two advanced pre-processing methods for multi-label disease classification on chest radiographs: bone suppression and lung field detection. In a systematic evaluation, we showed and discussed the superior performance of models -- trained on pre-processed images. The best performance was achieved by a novel ensemble architecture leveraging all the information from the different pre-processing methods. Significant AUC improvement for specific classes like \textit{foreign object} and \textit{mass} have been achieved, but there is still work needed for a clinical application. Our future work will include detailed investigation of clinical application scenarios and the integration of disease segmentation for multi-label classification.

	\printbibliography

\end{document}